\documentclass[10pt,twocolumn,letterpaper]{article}

\usepackage{cvpr}              % To produce the CAMERA-READY version
\usepackage{graphicx}
\usepackage{amsmath}
\usepackage{amssymb}
\usepackage{booktabs}
\usepackage{placeins} 
\usepackage{float} 
\usepackage{enumitem}
\usepackage{tabularx} 
\usepackage{longtable} 
\usepackage{array} 
\usepackage{ragged2e} 
\usepackage{booktabs} 
\usepackage{placeins}
\usepackage{afterpage} 
\usepackage{caption} 

\usepackage[pagebackref,breaklinks,colorlinks]{hyperref}

\usepackage[capitalize]{cleveref}
\crefname{section}{Sec.}{Secs.}
\Crefname{section}{Section}{Sections}
\Crefname{table}{Table}{Tables}
\crefname{table}{Tab.}{Tabs.}

\def\cvprPaperID{*} % *** Enter the CVPR Paper ID here
\def\confName{CVPR}
\def\confYear{2025}

\begin{document}

%%%%%%%%% TITLE
\title{DVI: Disentangling Semantic and Visual Identity for Training-Free Personalized Generation}

% \author{First Author\\
% Affiliation\\
% {\tt\small first.author@email.com}
% \and
% Second Author\\
% Affiliation\\
% {\tt\small second.author@email.com}
% }
\author{Guandong Li\\
iFLYTEK\\
 % \quad (Corresponding Author)\\
\and
Yijun Ding\\
Suning\\
}

\twocolumn[{
\renewcommand\twocolumn[1][]{#1}
\maketitle
% \begin{center}
%     \centering
%     \vspace*{-.8cm}
%     % Please ensure the image exists
%     \includegraphics[width=\textwidth]{pic/figure1.png} 
%     \vspace*{-.6cm}
%     \captionof{figure}{Overview of the DVI Framework. We decouple identity customization into a person ID branch (bottom) for extracting semantic structure and an implicit visual statistics branch (top) leveraging VAE latent variables to extract atmospheric context. These heterogeneous features are synergistically fused via parameter-free feature modulation and regulated by a dynamic scheduler within the MM-DiT backbone, thereby simultaneously establishing identity fidelity and visual consistency.}
% \label{fig:figure1}
% \end{center}
}]

%%%%%%%%% ABSTRACT
\begin{abstract}
Recent tuning-free identity customization methods have achieved significant success in maintaining facial fidelity by leveraging high-level semantic embeddings from pre-trained face recognition models. However, existing methods often overlook the visual context inherent in reference images—such as lighting distribution, skin texture density, and environmental tone. This limitation frequently leads to ``Semantic-Visual Dissonance,'' where the generated identity retains accurate facial geometry but loses the unique atmosphere and texture of the input image, resulting in an unnatural ``sticker-like'' effect. To bridge this gap, we propose DVI (Disentangled Visual-Identity), a novel zero-shot framework that orthogonally disentangles identity customization into a fine-grained semantic stream and a coarse-grained visual stream. Unlike methods relying solely on semantic vectors, DVI exploits the inherent statistical properties of the generative model's VAE latent space. We elucidate that the first and second-order statistics (mean and variance) of VAE latent variables serve as powerful yet lightweight descriptors for characterizing global visual atmosphere. To fuse these heterogeneous features, we propose a Parameter-Free Feature Modulation mechanism. Instead of training learnable projection layers, this mechanism adaptively modulates the distribution of high-dimensional semantic embeddings using low-dimensional visual statistics, effectively injecting the ``visual soul'' of the reference image into the generation process. Furthermore, we design a Dynamic Temporal Granularity Scheduler to align with the coarse-to-fine generation characteristic of diffusion models, prioritizing the injection of visual atmosphere in the early denoising stages while refining semantic details in the later stages. Extensive experiments demonstrate that DVI significantly enhances visual consistency and atmospheric fidelity without any parameter fine-tuning, while maintaining robust identity preservation, outperforming existing state-of-the-art methods and achieving excellent results in IBench evaluation.
\end{abstract}

%%%%%%%%% BODY TEXT
\section{Introduction}
\label{sec:intro}

Personalized text-to-image generation \cite{wang2024instantid, gal2022image, kumari2023multi, liu2023facechain, li2025editid, li2025editidv2,li2024layout,li2023smartbanner} aims to seamlessly integrate specific subject identities into user-provided text descriptions. This technology holds immense application potential in fields such as digital avatar creation, immersive storytelling, and virtual try-on. With the rapid development of diffusion models, the research paradigm in this field has gradually shifted from optimization-based methods (e.g., LoRA \cite{hu2022lora}, DreamBooth \cite{ruiz2023dreambooth}) to more efficient tuning-free methods \cite{guo2024pulid, mou2025dreamo, chen2025xverse, jiang2025infiniteyou}. Tuning-free methods achieve ``plug-and-play'' zero-shot generation by directly extracting identity features through feed-forward encoders, greatly enhancing user experience.

However, existing tuning-free paradigms face a long-overlooked challenge: \textit{The trade-off between semantic fidelity and visual consistency}. Current mainstream methods mainly rely on pre-trained face recognition models (e.g., ArcFace \cite{deng2019arcface}) or multi-modal encoders (e.g., CLIP \cite{radford2021learning}) to extract high-level semantic embeddings from reference images \cite{xiao2025fastcomposer, he2025uniportrait, li2024photomaker}. While these embeddings effectively capture the ``Who'' identity information, their highly compressed nature often discards low-level visual cues of ``In what context,'' such as the unique lighting distribution, film grain, skin texture density, and overall tonal atmosphere of the reference image.

The absence of visual context frequently leads to Semantic-Visual Dissonance in generated results: while the generated faces retain accurate geometric fidelity to the subject, their lighting, texture, and sharpness often appear generic and over-smoothed, disjointed from the unique visual style of the reference image. This phenomenon is particularly pronounced when processing reference images with strong artistic styles or complex lighting conditions, resulting in a ``pasted-on'' artifact where the subject lacks organic fusion with the background.

To address this issue, we revisit the composition of identity features. We argue that high-fidelity identity customization requires not only explicit semantic disentanglement but also implicit visual statistics transfer. Based on this insight, we propose \textbf{DVI (Disentangled Visual-Identity)}, a novel training-free identity injection framework.

The core idea of DVI is to orthogonally decompose a single identity feature stream into two complementary streams:
\begin{itemize}[leftmargin=*]
    \item \textbf{Fine-Grained Semantic Stream}: Following existing paradigms, it extracts high-dimensional semantic features to ensure precise reconstruction of facial geometry.
    \item \textbf{Coarse-Grained Visual Stream}: This is our core innovation. Instead of introducing additional heavy encoders, we exploit the potential of the generative model's own VAE (Variational Autoencoder) latent space. We find that the global first-order (mean) and second-order (variance) statistics of VAE latent feature maps can extremely efficiently characterize the overall visual atmosphere and texture distribution of an image.
\end{itemize}

% \begin{itemize}[leftmargin=*]
%     \item \textbf{Fine-Grained Semantic Stream}: Following existing paradigms, this stream extracts high-dimensional semantic features to ensure the precise reconstruction of facial geometry.
%     \item \textbf{Coarse-Grained Visual Stream}: This constitutes our core innovation. Instead of introducing computationally heavy encoders, we exploit the generative model's inherent VAE (Variational Autoencoder) latent space. We identify that the global first-order (mean) and second-order (variance) statistics of VAE latent feature maps efficiently characterize the overall visual atmosphere and texture distribution.
% \end{itemize}

To fuse these two heterogeneous feature streams, we discard the linear projectors requiring massive data training in traditional methods and design a Parameter-Free Feature Modulation mechanism. Inspired by the AdaIN concept in style transfer, we use visual statistics as a global distribution bias to directly modulate the distribution of semantic embeddings in the feature space. This approach not only avoids introducing any trainable parameters but also demonstrates strong robustness. Furthermore, considering the temporal characteristics of the generation process, we introduce a Dynamic Temporal Granularity Scheduler to adaptively balance the weights of visual atmosphere laying and semantic detail refinement across different generation stages.

In summary, the contributions of this paper are as follows:
\begin{enumerate}
    \item \textbf{Propose the DVI Framework}: We present a dual-granularity identity disentanglement framework that, for the first time in a tuning-free setting, explicitly distinguishes and synergistically utilizes the semantic and visual attributes of identity, solving the problem where generated images look like the person but lack the correct vibe (resembling in form but not in spirit).
    \item \textbf{Innovative Use of VAE Statistics}: We demonstrate that without complex attention mechanisms, simply extracting statistical quantities from the VAE latent space effectively captures and transfers the visual atmosphere and texture of the reference image.
    \item \textbf{Parameter-Free Modulation and Dynamic Scheduling}: We design a training-free feature modulation module and temporal scheduling strategy, significantly improving generation quality and identity consistency in complex lighting and narrative scenes with zero extra training cost.
\end{enumerate}

\section{Related Works}

\subsection{Tuning-Free Identity Customization}
With the popularity of diffusion models, personalized generation has evolved from optimization-based methods to more efficient tuning-free methods. These methods aim to achieve identity preservation directly through feed-forward processes without time-consuming test-time fine-tuning for specific subjects. Early attempts like IP-Adapter \cite{ye2023ip} utilized decoupled cross-attention mechanisms to introduce image prompts, achieving general style and content transfer but with limited fine-grained facial identity preservation. To enhance identity fidelity, recent SOTA methods (e.g., InstantID \cite{wang2024instantid}, PhotoMaker \cite{li2024photomaker}, FaceClip \cite{liu2025learning}) typically introduce powerful face recognition models (e.g., ArcFace \cite{deng2019arcface}) as feature extractors. These methods compress reference images into highly abstract semantic embeddings and inject them into the UNet or Transformer of diffusion models by training linear projection layers.

Although these methods have achieved significant results in Identity Recognition, their highly compressed feature representations often lose low-level visual cues from reference images (e.g., skin details, lighting patterns, and environmental tones). This Semantic-Visual Imbalance results in generated images often presenting a homogenized visual effect lacking the original atmosphere. In contrast, DVI is dedicated to completing this missing visual context link without requiring training.

\subsection{Visual Context Preservation}
Beyond identity customization, preserving the visual attributes (style, texture, layout) of reference images has always been a key topic in generative models. Traditional Style Transfer methods typically use the Gram matrix of Convolutional Neural Networks to capture texture statistical information. In the era of diffusion models, ControlNet \cite{zhang2023adding} and T2I-Adapter \cite{mou2024t2i} introduce extra control branches to preserve spatial structural information, but they focus more on geometric contours than texture atmosphere. Some recent studies have begun to explore using the latent space of Variational Autoencoders (VAE) to preserve visual consistency \cite{wu2025less, cheng2025umo, mao2024realcustom++, mou2025dreamo}. Unlike semantic encoders like CLIP, VAE latent variables retain the spatial layout and pixel-level statistical properties of images. Although some works attempt to align VAE features by training complex Reference Attention mechanisms, this introduces high computational and training costs.

DVI proposes a lighter perspective: we believe the ``Visual Vibe'' of an image can be effectively characterized by the Global Mean and Variance of VAE latent feature maps. This discovery allows us to bypass heavy attention mechanism calculations and directly use statistics as visual descriptors to achieve zero-shot visual atmosphere transfer.

\subsection{Feature Modulation and Fusion}
Injecting external conditions into diffusion models typically involves two paradigms: Cross-Attention and Concatenation. However, when fusing heterogeneous features (e.g., high-dimensional semantic features and low-dimensional visual statistics), direct concatenation often leads to feature space misalignment. Existing solutions usually rely on training learnable MLP projection layers to map feature dimensions, which not only increases model parameters but may also lead to overfitting or damaging pre-trained model priors. Inspired by Adaptive Instance Normalization (AdaIN \cite{gu2021adain}) in style generation (StyleGAN \cite{karras2019style}), we explore the application of Parameter-Free Feature Modulation in diffusion models. Unlike modules requiring training, DVI uses visual statistics to directly modulate the distribution (shift and scale) of semantic features. This method avoids extra training costs and achieves orthogonal disentanglement and organic fusion of semantic and visual streams in the feature space.

\section{Method}

The core objective of DVI is to orthogonally disentangle the Fine-Grained Semantic Identity and Coarse-Grained Visual Context from the reference image $I_{ref}$ and synergistically inject them into a pre-trained DiT (Diffusion Transformer) generative model under Zero-Shot and Tuning-Free constraints.

\begin{figure*}[t]
  \centering
  \includegraphics[width=\textwidth]{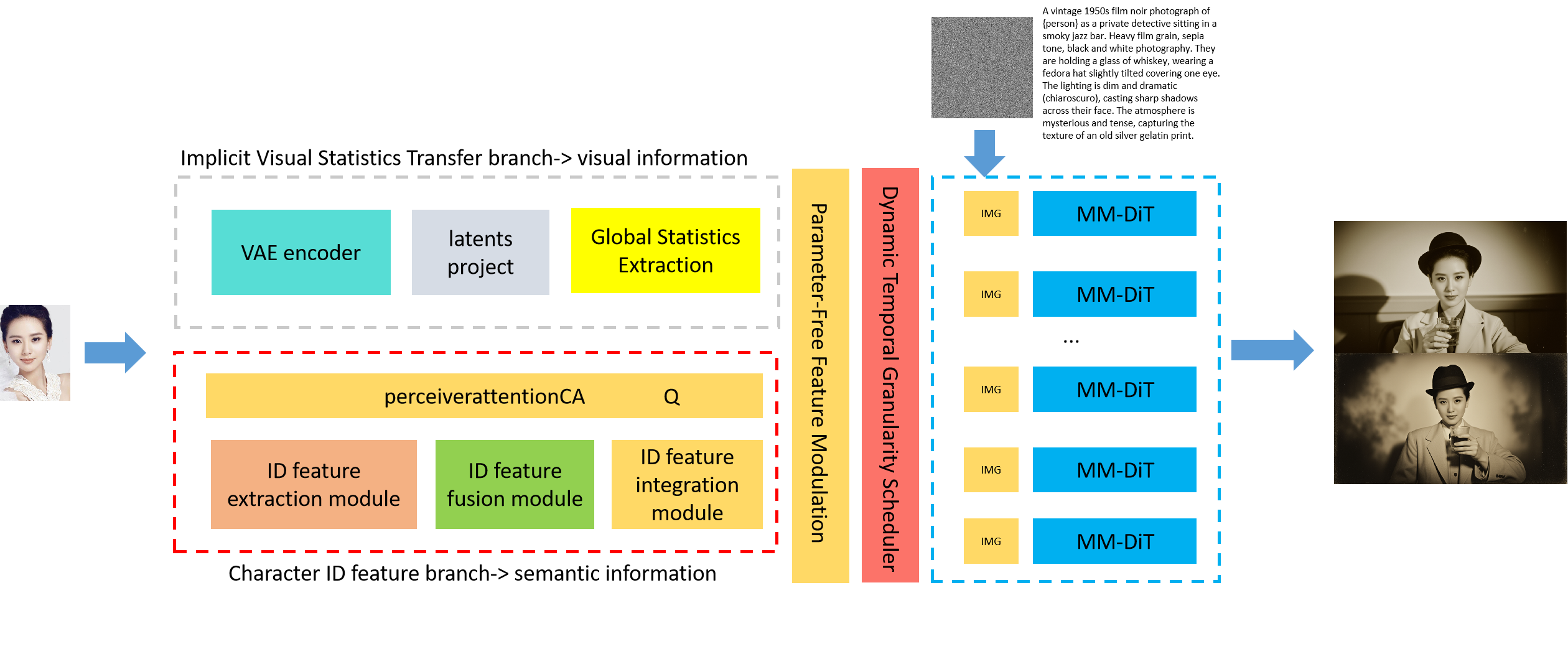} 
  \caption{Overview of the DVI Framework. We disentangle identity customization into a person ID branch (bottom) for extracting semantic structure and an implicit visual statistics branch (top) leveraging VAE latent variables to extract atmospheric context. These heterogeneous features are synergistically fused via parameter-free feature modulation and regulated by a dynamic scheduler within the MM-DiT backbone.}
  \label{fig:method}
\end{figure*}

\subsection{Coarse-Grained Visual Stream}
Existing ID customization methods often employ aggressive image preprocessing, which destroys the compositional proportions and texture density of original images. To capture the global visual atmosphere (e.g., lighting distribution, tone, and film texture) of the reference image, we construct the Coarse-Grained Visual Stream.

\paragraph{Preserve-Aspect-Ratio Latent Encoding.}
First, we define a texture density preserving preprocessing strategy $\mathcal{P}$. Given a reference image $I_{ref}$, we first scale it proportionally based on its shortest edge, followed by a center crop to obtain a fixed-resolution input $x_{vis} \in \mathbb{R}^{3 \times H \times W}$.
\begin{equation}
x_{vis} = \mathcal{P}(I_{ref})
\end{equation}
Subsequently, we use the encoder $\mathcal{E}$ of a frozen pre-trained Variational Autoencoder (VAE) to map $x_{vis}$ to the latent space, obtaining the latent feature map $Z$:
\begin{equation}
Z = \mathcal{E}(x_{vis}) \in \mathbb{R}^{C \times h \times w}
\end{equation}
where $C$ is the number of latent channels (in our setting $C=16$), and $h, w$ are the spatial dimensions after downsampling.

\paragraph{Global Statistics Extraction.}
We assume that the ``visual atmosphere'' of an image can be characterized by the statistical distribution of latent features across channel dimensions. Unlike semantic features that focus on specific spatial structures, visual atmosphere is closer to a global style distribution. Therefore, we calculate the first-order moment (mean $\mu$) and second-order moment (standard deviation $\sigma$) of $Z$ in the spatial dimension:

\begin{equation}
\mu_c = \frac{1}{h \cdot w} \sum_{i=1}^{h} \sum_{j=1}^{w} Z_{c,i,j}
\end{equation}

\begin{equation}
\sigma_c = \sqrt{\frac{1}{h \cdot w} \sum_{i=1}^{h} \sum_{j=1}^{w} (Z_{c,i,j} - \mu_c)^2 + \epsilon}
\end{equation}

Finally, we concatenate the mean and standard deviation to form a compact visual context descriptor $v_{ctx}$:
\begin{equation}
v_{ctx} = \text{Concat}(\mu, \sigma) \in \mathbb{R}^{2C}
\end{equation}
This descriptor (dimension 32) efficiently encodes the brightness, contrast, and texture statistical properties of the reference image, with negligible computational cost.

\subsection{Fine-Grained Semantic Stream}
To maximize editability while ensuring high-fidelity identity consistency, we do not simply treat identity features as a static vector but construct a hierarchical Semantic Stream Pipeline. Referring to advanced ID customization architectures, we further decouple this stream into three tightly coupled sub-modules.

\paragraph{ID Feature Extraction Module.}
This module aims to capture multi-level identity information from the reference image $I_{ref}$. We use a pre-trained Face Recognition Backbone as the encoder $\mathcal{S}$. To balance overall identity recognition and local facial geometric details, we extract Local Features from the intermediate layers and Global Features from the top layer of the network.
\begin{equation}
F_{raw} = \mathcal{S}(I_{ref}) = \{f_{global}, f_{local}\}
\end{equation}
These raw features form the ``semantic base'' of the identity, covering facial topological structures but not yet aligned with the generative model's text-image space.

\paragraph{ID Feature Fusion Module.}
To address the heterogeneity between the ID feature space and the diffusion model's Latent Space, and since direct injection of raw features leads to reduced editability (i.e., ``sticker effect''), we design a fusion projection network. This module maps $F_{raw}$ to a semantic space aligned with the text encoder, producing Mapped Features.
In this process, we introduce the concept of Offset Features, utilizing a learnable residual term to fine-tune feature direction, granting flexibility to adapt to different contexts while maintaining core ID attributes.
\begin{equation}
f_{id} = \text{Proj}(F_{raw}) + \delta_{offset} \in \mathbb{R}^{N \times D}
\end{equation}
where $f_{id}$ is the final aligned high-dimensional semantic embedding (in our setting $D=2048$). This step is crucial as it ensures the ID features are ``editable'' rather than rigid pixel copies.

\paragraph{ID Feature Insertion Module.}
Finally, we inject $f_{id}$ into the generation backbone of the DiT via Cross-Attention. To balance ID preservation and responsiveness to text instructions, we adopt a Dynamic Embedding Strength design. Unlike traditional fixed-weight injection, we dynamically adjust the injection strength $\alpha_l$ at different layers $l$ of the DiT:
\begin{equation}
\text{Attn}(Q, K, V) = \text{Softmax}\left(\frac{Q K^T}{\sqrt{d}}\right) (V + \alpha_l \cdot f_{id})
\end{equation}
This design constitutes the ``semantic anchor'' of DVI. Notably, at this point $f_{id}$ possesses perfect geometric structure and identity semantics but still lacks specific visual atmosphere. This is precisely where we introduce the Coarse-Grained Visual Stream for modulation in the next section.

\subsection{Parameter-Free Feature Modulation (PFFM)}
This is the core innovation of DVI. The challenge lies in fusing the low-dimensional visual statistics $v_{ctx}$ (32-dim) into the high-dimensional semantic embeddings $f_{id}$ (2048-dim). Traditional approaches train a Multi-Layer Perceptron (MLP) for dimensional projection, but this violates the Tuning-Free premise.

Inspired by the principle of Adaptive Instance Normalization (AdaIN), we propose a Parameter-Free Feature Modulation mechanism. We treat visual statistics as a Global Distribution Bias to dynamically adjust the distribution of semantic embeddings in the feature space.

\paragraph{Dimension Broadcasting.}
First, since the dimension of $v_{ctx}$ is much smaller than $D$, we extend it to the target dimension $D$ via a Repeat/Tile operation, constructing the modulation vector $m_{vis}$:
\begin{equation}
m_{vis} = \text{Repeat}(v_{ctx}, \lceil D / 2C \rceil)[:D]
\end{equation}

\paragraph{Distribution Injection.}
To inject visual atmosphere into semantic features, we perform a normalized modulation operation on $f_{id}$. Unlike AdaIN which requires learning scale factors $\gamma$ and offset factors $\beta$, we directly use $m_{vis}$ as an additive bias to offset the manifold distribution of semantic features:

\begin{equation}
f_{fused} = \text{Norm}(f_{id}) + \lambda(t) \cdot \Psi(m_{vis})
\end{equation}

where $\text{Norm}(\cdot)$ denotes Layer Normalization, $\Psi$ is a simple scaling operator (coefficient set to 0.5 in implementation), and $\lambda(t)$ is a time-variant control coefficient.
The physical meaning of this operation is: $f_{id}$ determines the ``center position'' of features in the semantic space (i.e., who the identity is), while $m_{vis}$ provides a ``directional offset'' to this center pointing towards a specific visual style (i.e., in what lighting atmosphere).

\subsection{Dynamic Temporal Granularity Scheduler}
The diffusion generation process is inherently a Coarse-to-Fine denoising process: early time steps mainly determine global composition, tone, and lighting, while later time steps focus on refining local textures and facial details.

If we inject visual features with constant strength throughout the generation process, it might lead to excessive stylization destroying facial details. Therefore, we design a Dynamic Temporal Granularity Scheduler. We define the visual modulation strength weight $\lambda(t)$ to decay linearly with the denoising time step $t$ (from 1.0 noise state to 0.0 clean state):

\begin{equation}
\lambda(t) = \lambda_{base} \cdot t
\end{equation}

\begin{itemize}[leftmargin=*]
    \item \textbf{Phase 1 (High Noise, $t \to 1.0$)}: $\lambda(t)$ is large. Visual stream dominates; the model uses VAE statistics to lay down the overall tone and compositional atmosphere.
    \item \textbf{Phase 2 (Low Noise, $t \to 0.0$)}: $\lambda(t)$ approaches 0. Semantic stream dominates; the model focuses on using $f_{id}$ to refine facial details, ensuring identity fidelity.
\end{itemize}

This dynamic balancing strategy ensures the organic coexistence of visual atmosphere and identity semantics in the temporal dimension, avoiding feature conflicts.

\section{Experiments}
\label{sec:experiments}

\subsection{Experimental Settings}
\paragraph{Implementation Details.}
Our method is implemented based on the Flux.1 model. For the Fine-Grained Semantic Stream, we employ AntelopeV2 as the face recognition backbone and CLIP ViT-L/14 as the auxiliary image encoder, reusing the feature embedding and feature fusion module weights from PuLID. For the Coarse-Grained Visual Stream, we directly reuse Flux's native VAE Encoder without loading any additional models.
In the inference stage, the default sampling steps are set to $T=25$, guidance to 4, and the sampler uses Euler. To leave sufficient feature modulation space for the visual stream, we set the ID injection weight to 0.8 (slightly lower than the Baseline's 1.0). All experiments are completed on 4 NVIDIA A100 (80GB) GPUs. Our evaluation adopts the IBench framework proposed by EditID\cite{li2025editid}.

\subsection{Qualitative Comparison}
\begin{figure*}[t]
    \centering
    \includegraphics[width=\textwidth]{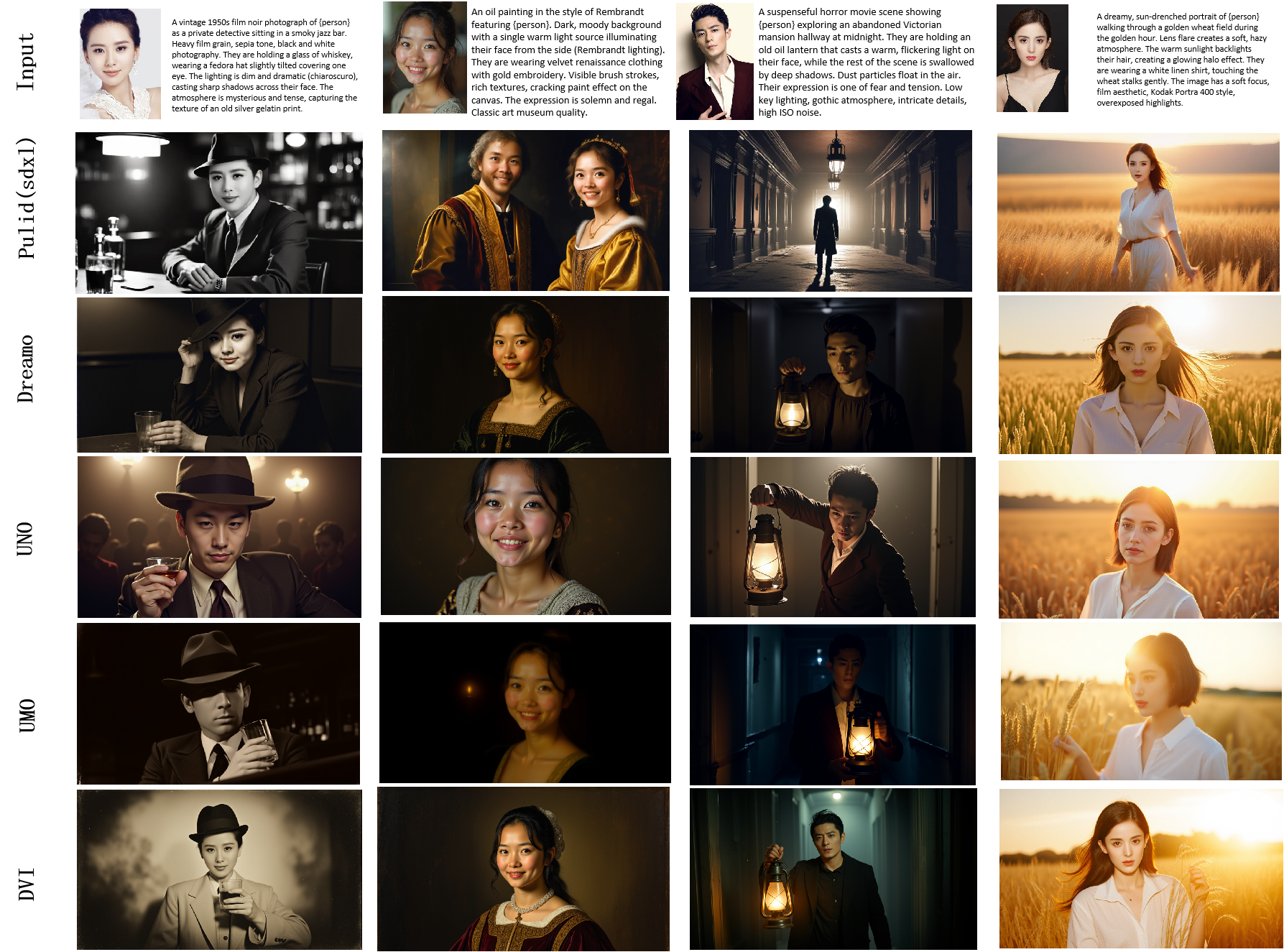}
    \caption{Qualitative Comparison: DVI achieves higher editability while ensuring ID consistency. DVI accurately realizes character consistency and complete presentation of atmospheric visual concepts in complex narrative scenes.}
    \label{fig:qualitative_comparison}
\end{figure*}

We compare DVI with PuLID (SDXL version \cite{podell2023sdxl}), PuLID (Flux Krea version), DreamO \cite{mou2025dreamo}, UNO \cite{wu2025less}, and UMO \cite{cheng2025umo}. The base model for PuLID (SDXL) is SDXL\_base\_1.0, while all other models (including DVI) use Flux.1 dev. As shown in Figure \ref{fig:qualitative_comparison}, we input long text prompts containing strong stylistic descriptions. While maintaining character consistency, DVI achieves superior Visual Atmosphere Integration compared to PuLID; compared to UNO and DreamO, DVI not only avoids severe ID loss but also generates images with more natural texture and light-shadow interaction, strictly adhering to environmental instructions in the prompt.

Figure \ref{fig:qualitative_comparison} displays four challenging stylized scenes: vintage film noir detective, classical oil painting portrait, horror suspense corridor, and backlit pastoral memory. In the first column (vintage film), the prompt explicitly requests ``Heavy film grain, sepia tone... texture of an old silver gelatin print''. PuLID (SDXL) and DreamO generate faces that are too clean and sharp, presenting a jarring modern digital photo look forcibly desaturated, completely disjointed from the background's grainy texture. In contrast, DVI successfully ``injects'' film grain into the facial skin, presenting a natural yellowish aged look, perfectly fitting the ``old photo'' narrative atmosphere. In the second column (classical oil painting), facing the description ``Visible brush strokes, rich textures, cracking paint effect'', UNO generates a painting style but with severe identity drift. DVI, while precisely maintaining ID, renders facial brush stroke textures using the impasto method, appearing as if painted on canvas rather than a pasted photo.

In the third column (horror corridor) and fourth column (backlit pastoral), light-shadow interaction is key. In the third column, DreamO's facial lighting is flat, failing to reflect the warm light source characteristics of the lantern; in the fourth column, PuLID's face retains the white light of the original image without producing the ``glowing halo effect'' under backlight. DVI shows excellent lighting adaptability in both scenes: the face is naturally illuminated by warm candlelight in the horror scene, and shows rim light in the pastoral scene, demonstrating DVI's precise capture of Visual Context and flexible reorganization of ID features in complex scenes with strong environmental light interference.

We further verified the above visual observations using metrics from the ChineseID + Editable Long Prompts benchmark, as shown in Table \ref{tab:chineseid_evaluation}. DVI surpasses existing SOTA methods on multiple key metrics.
On Aesthetic Quality \& Visual Atmosphere, DVI achieves the highest scores in Aesthetic (0.700) and Image Quality (0.515) among all compared models. Compared to SDXL-based PuLID (0.675) and the latest PuLID-Krea (0.683) and DreamO (0.678), DVI's significant improvement strongly proves the effectiveness of the Coarse-Grained Visual Stream. By introducing VAE latent space statistics, DVI successfully eliminates the ``Semantic-Visual Dissonance'' common in traditional ID injection methods.
On the trade-off between Identity Preservation vs. Editability, DVI reaches 0.557 on the FaceSim metric, ranking first and significantly outperforming PuLID-Krea (0.495) and DreamO (0.398). This indicates DVI's semantic stream builds the most solid identity anchor. More importantly, while maintaining high ID fidelity, DVI retains extremely high Exprdiv (0.601), on par with UNO (0.614) which has strong style transfer but severe ID loss (FaceSim 0.105). This reflects DVI's position at a ``Golden Balance Point'': it does not sacrifice ID for editability like UNO, nor sacrifice expression flexibility for ID like traditional models.
On Visual \& Textual Consistency, DVI achieves 0.804 on ClipI, almost tying with the best DreamO (0.805), but DVI's ID preservation capability (0.557) far exceeds DreamO (0.398). This indicates DVI's PFFM mechanism successfully injects the ``visual soul'' of the reference image. Additionally, the high ClipT score (0.269) proves DVI does not sacrifice adherence to text prompts by introducing the visual stream.

\begin{table*}[t]
    \centering
    \caption{Evaluation metric results from IBench on ChineseID with editable long prompts}
    \label{tab:chineseid_evaluation}
    \begin{tabular}{lcccccc}
        \toprule
        Model & Aesthetic & Image Quality & Exprdiv & Facesim & ClipI & ClipT \\
        \midrule
        PuLID (SDXL) & 0.675 & 0.502 & 0.593 & 0.399 & 0.768 & 0.248 \\
        PuLID (Krea) & 0.683 & 0.505 & 0.587 & 0.495 & 0.793 & 0.277 \\
        DreamO & 0.678 & 0.510 & 0.601 & 0.398 & 0.805 & 0.266 \\
        UNO & 0.675 & 0.465 & 0.614 & 0.105 & 0.797 & 0.267 \\
        UMO & 0.669 & 0.469 & 0.619 & 0.397 & 0.748 & 0.259 \\
        DVI (Ours) & \textbf{0.700} & \textbf{0.515} & 0.601 & \textbf{0.557} & 0.804 & 0.269 \\
        \bottomrule
    \end{tabular}
\end{table*}

\subsection{Ablation Study}

\subsubsection{Effectiveness of Coarse-Grained Visual Stream}
To verify the key role of the Coarse-Grained Visual Stream in introducing style and atmospheric elements, we compared the full DVI model with a version removing the visual stream module.

\begin{figure}[t]
\centering
\includegraphics[width=\linewidth]{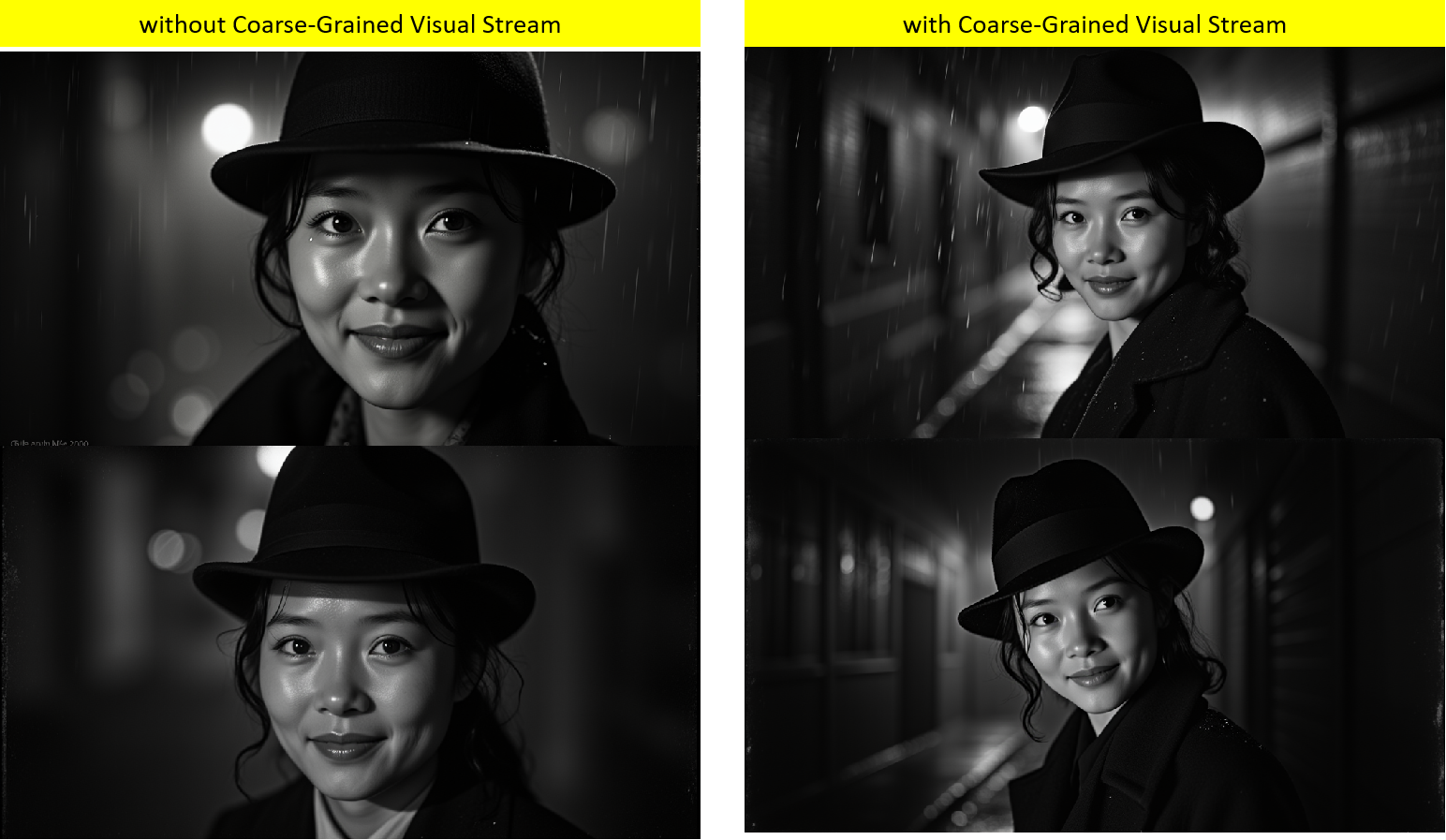}
\caption{Comparison between removing the visual stream (Left) and the full DVI model (Right). The left image retains ID but the face is too independent and clean, showing obvious Foreground-Background Detachment; the right image successfully injects atmospheric elements like film grain and low-key lighting via the visual stream.}
\label{fig:ablation_visual}
\end{figure}

\textbf{Foreground-Background Detachment Phenomenon}: When the visual stream is removed (as shown in Figure \ref{fig:ablation_visual} Left), the model relies solely on the fine-grained semantic stream. Although facial geometric structures are accurately restored, the face presents a ``clarity ignoring the environment''. In a Film Noir scene emphasizing low light and grain, the face in the left image retains the high signal-to-noise ratio and uniform lighting of a modern digital photo, clashing with the dim, rough rainy night background.

\textbf{Injection of Visual Atmosphere Elements}: In contrast, the full scheme introducing the coarse-grained visual stream (Figure \ref{fig:ablation_visual} Right) successfully ``injects'' the style elements of the reference image into the identity features. Through VAE statistics modulation, the character's face is no longer an independent semantic island but is endowed with more environmental atmospheric elements: facial lighting ratios are increased to match the oppressive feel of film noir, skin texture becomes rough to echo film grain, and edge contours soften to blend into the rainy mist.

\subsubsection{Comparison of Feature Modulation Designs}
\begin{figure}[t]
\centering
\includegraphics[width=\linewidth]{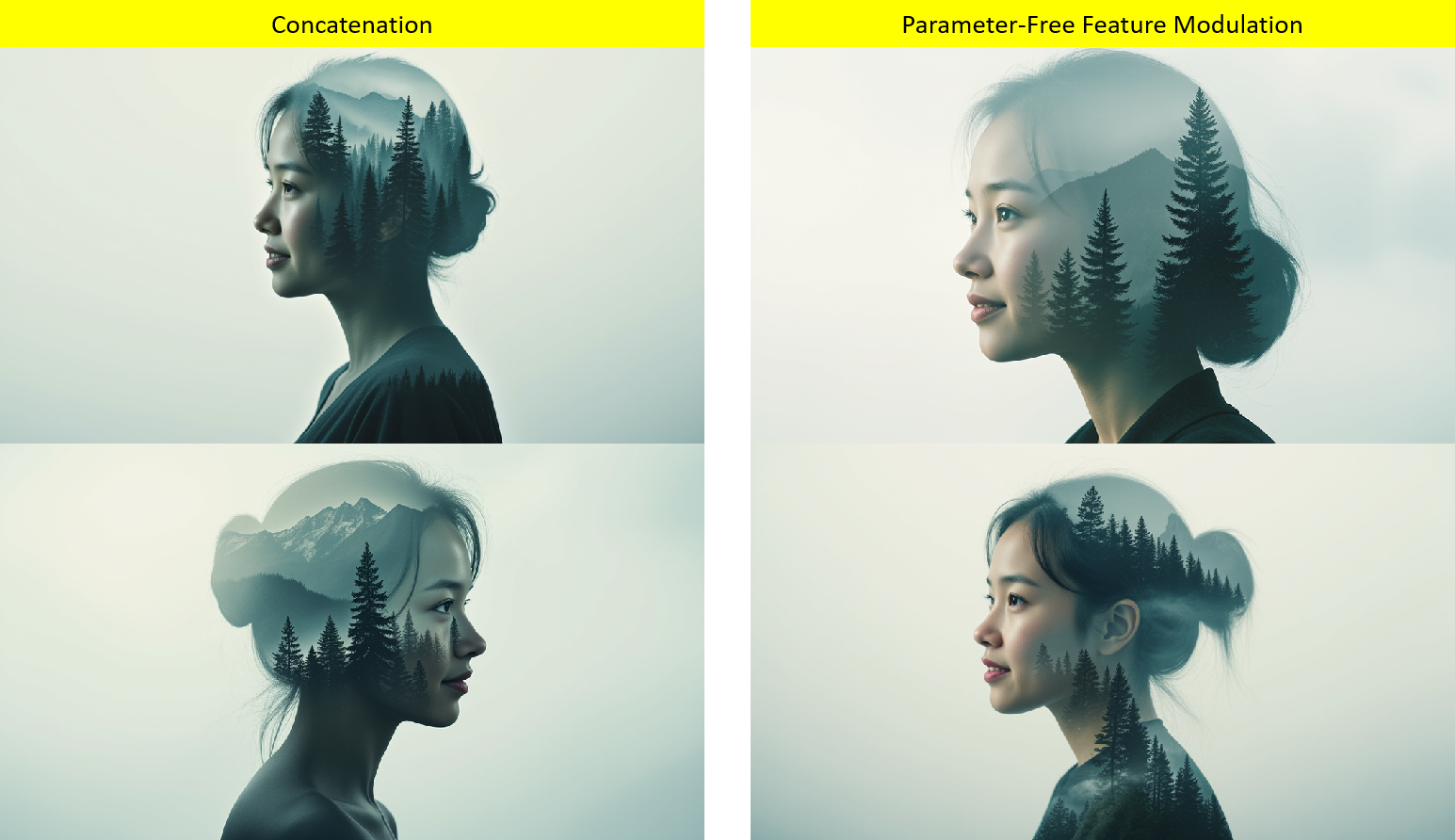}
\caption{Comparison between simple feature Concatenation (Left) and DVI Parameter-Free Feature Modulation (PFFM, Right) under Double Exposure style. Concatenation leads to muddy tones and rigid edges; PFFM achieves smooth natural fusion.}
\label{fig:ablation_concat}
\end{figure}

We further compared the effects of Parameter-Free Feature Modulation (PFFM) versus simple Concatenation strategies.
Under the Tuning-Free setting, simple concatenation often leads to dimensional abruptness in feature space, making it difficult for the pre-trained DiT model to adapt to the forced inclusion of heterogeneous features, often manifesting as ``Distribution Mismatch''. This is particularly evident when handling ``Double Exposure'' art styles requiring high structural fusion.
As shown in Figure \ref{fig:ablation_concat} Left, images generated using the concatenation strategy show rigid structural conflicts at the junction of the portrait silhouette and the landscape (forest). Facial tones appear ``Muddy Tones'' and covered by shadow, and the transition between hair and trees lacks naturalness.
In contrast, DVI's distribution modulation strategy based on AdaIN demonstrates high robustness. This mechanism uses visual statistics as a global bias to dynamically shift and scale the distribution of semantic features. As shown in Figure \ref{fig:ablation_concat} Right, PFFM achieves dreamlike ``Organic Fusion''.

\section{Conclusion}
This paper proposes DVI, a zero-shot identity customization framework for text-to-image generation. Addressing the ``Semantic-Visual Dissonance'' caused by excessive reliance on semantic feature compression in existing Tuning-Free methods, we provide an elegant training-free solution. The core contributions of DVI lie in reformulating the identity customization task as the synergistic injection of dual-granularity features: 1. Constructing a hierarchical semantic stream to provide a solid identity geometric anchor and editability foundation. 2. Mining the inherent potential of the generative model's latent space, using VAE statistics (mean and variance) as lightweight global visual descriptors. 3. Designing Parameter-Free Feature Modulation (PFFM) and Dynamic Temporal Scheduling to successfully inject the ``visual soul'' (lighting, texture, atmosphere) of the reference image into semantic features without extra training costs. Experimental results show that DVI maintains robust identity consistency while significantly enhancing artistic beauty and environmental integration in high-complexity narrative scenes.

%%%%%%%%% REFERENCES
{\small
\bibliographystyle{ieee_fullname}
\bibliography{egbib}

@String(ICLR = {Int. Conf. Learn. Represent.})

@String(AAAI = {AAAI})

@String(ICLR  = {ICLR})

@article{wang2024instantid,
  title={Instantid: Zero-shot identity-preserving generation in seconds},
  author={Wang, Qixun and Bai, Xu and Wang, Haofan and Qin, Zekui and Chen, Anthony and Li, Huaxia and Tang, Xu and Hu, Yao},
  journal={arXiv preprint arXiv:2401.07519},
  year={2024}
}

@article{gal2022image,
  title={An image is worth one word: Personalizing text-to-image generation using textual inversion},
  author={Gal, Rinon and Alaluf, Yuval and Atzmon, Yuval and Patashnik, Or and Bermano, Amit H and Chechik, Gal and Cohen-Or, Daniel},
  journal={arXiv preprint arXiv:2208.01618},
  year={2022}
}

@inproceedings{kumari2023multi,
  title={Multi-concept customization of text-to-image diffusion},
  author={Kumari, Nupur and Zhang, Bingliang and Zhang, Richard and Shechtman, Eli and Zhu, Jun-Yan},
  booktitle={Proceedings of the IEEE/CVF conference on computer vision and pattern recognition},
  pages={1931--1941},
  year={2023}
}

@inproceedings{ruiz2023dreambooth,
  title={Dreambooth: Fine tuning text-to-image diffusion models for subject-driven generation},
  author={Ruiz, Nataniel and Li, Yuanzhen and Jampani, Varun and Pritch, Yael and Rubinstein, Michael and Aberman, Kfir},
  booktitle={Proceedings of the IEEE/CVF conference on computer vision and pattern recognition},
  pages={22500--22510},
  year={2023}
}

@article{hu2022lora,
  title={Lora: Low-rank adaptation of large language models.},
  author={Hu, Edward J and Shen, Yelong and Wallis, Phillip and Allen-Zhu, Zeyuan and Li, Yuanzhi and Wang, Shean and Wang, Lu and Chen, Weizhu and others},
  journal={ICLR},
  volume={1},
  number={2},
  pages={3},
  year={2022}
}

@article{guo2024pulid,
  title={Pulid: Pure and lightning id customization via contrastive alignment},
  author={Guo, Zinan and Wu, Yanze and Zhuowei, Chen and Zhang, Peng and He, Qian and others},
  journal={Advances in neural information processing systems},
  volume={37},
  pages={36777--36804},
  year={2024}
}

@article{mou2025dreamo,
  title={Dreamo: A unified framework for image customization},
  author={Mou, Chong and Wu, Yanze and Wu, Wenxu and Guo, Zinan and Zhang, Pengze and Cheng, Yufeng and Luo, Yiming and Ding, Fei and Zhang, Shiwen and Li, Xinghui and others},
  journal={arXiv preprint arXiv:2504.16915},
  year={2025}
}

@article{chen2025xverse,
  title={XVerse: Consistent Multi-Subject Control of Identity and Semantic Attributes via DiT Modulation},
  author={Chen, Bowen and Zhao, Mengyi and Sun, Haomiao and Chen, Li and Wang, Xu and Du, Kang and Wu, Xinglong},
  journal={arXiv preprint arXiv:2506.21416},
  year={2025}
}

@article{jiang2025infiniteyou,
  title={InfiniteYou: Flexible photo recrafting while preserving your identity},
  author={Jiang, Liming and Yan, Qing and Jia, Yumin and Liu, Zichuan and Kang, Hao and Lu, Xin},
  journal={arXiv preprint arXiv:2503.16418},
  year={2025}
}

@inproceedings{deng2019arcface,
  title={Arcface: Additive angular margin loss for deep face recognition},
  author={Deng, Jiankang and Guo, Jia and Xue, Niannan and Zafeiriou, Stefanos},
  booktitle={Proceedings of the IEEE/CVF conference on computer vision and pattern recognition},
  pages={4690--4699},
  year={2019}
}

@inproceedings{radford2021learning,
  title={Learning transferable visual models from natural language supervision},
  author={Radford, Alec and Kim, Jong Wook and Hallacy, Chris and Ramesh, Aditya and Goh, Gabriel and Agarwal, Sandhini and Sastry, Girish and Askell, Amanda and Mishkin, Pamela and Clark, Jack and others},
  booktitle={International conference on machine learning},
  pages={8748--8763},
  year={2021},
  organization={PmLR}
}

@article{xiao2025fastcomposer,
  title={Fastcomposer: Tuning-free multi-subject image generation with localized attention},
  author={Xiao, Guangxuan and Yin, Tianwei and Freeman, William T and Durand, Fr{\'e}do and Han, Song},
  journal={International Journal of Computer Vision},
  volume={133},
  number={3},
  pages={1175--1194},
  year={2025},
  publisher={Springer}
}

@inproceedings{he2025uniportrait,
  title={Uniportrait: A unified framework for identity-preserving single-and multi-human image personalization},
  author={He, Junjie and Geng, Yifeng and Bo, Liefeng},
  booktitle={Proceedings of the IEEE/CVF International Conference on Computer Vision},
  pages={14399--14408},
  year={2025}
}

@inproceedings{li2024photomaker,
  title={Photomaker: Customizing realistic human photos via stacked id embedding},
  author={Li, Zhen and Cao, Mingdeng and Wang, Xintao and Qi, Zhongang and Cheng, Ming-Ming and Shan, Ying},
  booktitle={Proceedings of the IEEE/CVF conference on computer vision and pattern recognition},
  pages={8640--8650},
  year={2024}
}

@article{ye2023ip,
  title={Ip-adapter: Text compatible image prompt adapter for text-to-image diffusion models},
  author={Ye, Hu and Zhang, Jun and Liu, Sibo and Han, Xiao and Yang, Wei},
  journal={arXiv preprint arXiv:2308.06721},
  year={2023}
}

@article{liu2025learning,
  title={Learning Joint ID-Textual Representation for ID-Preserving Image Synthesis},
  author={Liu, Zichuan and Jiang, Liming and Yan, Qing and Jia, Yumin and Kang, Hao and Lu, Xin},
  journal={arXiv preprint arXiv:2504.14202},
  year={2025}
}

@article{li2025editid,
  title={EditID: Training-Free Editable ID Customization for Text-to-Image Generation},
  author={Li, Guandong and Chu, Zhaobin},
  journal={arXiv preprint arXiv:2503.12526},
  year={2025}
}

@article{li2025editidv2,
  title={EditIDv2: Editable ID Customization with Data-Lubricated ID Feature Integration for Text-to-Image Generation},
  author={Li, Guandong and Chu, Zhaobin},
  journal={arXiv preprint arXiv:2509.05659},
  year={2025}
}

@inproceedings{zhang2023adding,
  title={Adding conditional control to text-to-image diffusion models},
  author={Zhang, Lvmin and Rao, Anyi and Agrawala, Maneesh},
  booktitle={Proceedings of the IEEE/CVF international conference on computer vision},
  pages={3836--3847},
  year={2023}
}

@inproceedings{mou2024t2i,
  title={T2i-adapter: Learning adapters to dig out more controllable ability for text-to-image diffusion models},
  author={Mou, Chong and Wang, Xintao and Xie, Liangbin and Wu, Yanze and Zhang, Jian and Qi, Zhongang and Shan, Ying},
  booktitle={Proceedings of the AAAI conference on artificial intelligence},
  volume={38},
  number={5},
  pages={4296--4304},
  year={2024}
}

@article{wu2025less,
  title={Less-to-more generalization: Unlocking more controllability by in-context generation},
  author={Wu, Shaojin and Huang, Mengqi and Wu, Wenxu and Cheng, Yufeng and Ding, Fei and He, Qian},
  journal={arXiv preprint arXiv:2504.02160},
  year={2025}
}

@article{cheng2025umo,
  title={UMO: Scaling Multi-Identity Consistency for Image Customization via Matching Reward},
  author={Cheng, Yufeng and Wu, Wenxu and Wu, Shaojin and Huang, Mengqi and Ding, Fei and He, Qian},
  journal={arXiv preprint arXiv:2509.06818},
  year={2025}
}

@article{mao2024realcustom++,
  title={Realcustom++: Representing images as real-word for real-time customization},
  author={Mao, Zhendong and Huang, Mengqi and Ding, Fei and Liu, Mingcong and He, Qian and Zhang, Yongdong},
  journal={arXiv preprint arXiv:2408.09744},
  year={2024}
}

@inproceedings{karras2019style,
  title={A style-based generator architecture for generative adversarial networks},
  author={Karras, Tero and Laine, Samuli and Aila, Timo},
  booktitle={Proceedings of the IEEE/CVF conference on computer vision and pattern recognition},
  pages={4401--4410},
  year={2019}
}

@article{gu2021adain,
  title={AdaIN-based tunable CycleGAN for efficient unsupervised low-dose CT denoising},
  author={Gu, Jawook and Ye, Jong Chul},
  journal={IEEE Transactions on Computational Imaging},
  volume={7},
  pages={73--85},
  year={2021},
  publisher={IEEE}
}

@article{podell2023sdxl,
  title={Sdxl: Improving latent diffusion models for high-resolution image synthesis},
  author={Podell, Dustin and English, Zion and Lacey, Kyle and Blattmann, Andreas and Dockhorn, Tim and M{\"u}ller, Jonas and Penna, Joe and Rombach, Robin},
  journal={arXiv preprint arXiv:2307.01952},
  year={2023}
}

@article{liu2023facechain,
  title={Facechain: A playground for human-centric artificial intelligence generated content},
  author={Liu, Yang and Yu, Cheng and Shang, Lei and He, Yongyi and Wu, Ziheng and Wang, Xingjun and Xu, Chao and Xie, Haoyu and Wang, Weida and Zhao, Yuze and others},
  journal={arXiv preprint arXiv:2308.14256},
  year={2023}
}

@article{li2024layout,
  title={Layout Control and Semantic Guidance with Attention Loss Backward for T2I Diffusion Model},
  author={Li, Guandong},
  journal={arXiv preprint arXiv:2411.06692},
  year={2024}
}

@article{li2023smartbanner,
  title={Smartbanner: intelligent banner design framework that strikes a balance between creative freedom and design rules},
  author={Li, Guandong and Yang, Xian},
  journal={Multimedia Tools and Applications},
  volume={82},
  number={12},
  pages={18653--18667},
  year={2023},
  publisher={Springer}
}
}

\end{document}